\documentclass[sigconf]{acmart}
\usepackage{amsmath,amsfonts}
\usepackage{booktabs}
\usepackage{algorithmic}
\usepackage{graphicx}
\usepackage{textcomp}
\usepackage{xcolor}
\usepackage{url}
\usepackage{array}
\usepackage{multirow}
\usepackage{etoolbox}
\usepackage[most]{tcolorbox}
\usepackage{subfigure}
\usepackage{cases}
\usepackage{makecell}
\usepackage{color, soul}
\usepackage{bbding}
\usepackage{colortbl}
\usepackage{float}
\usepackage{arydshln}
\usepackage{placeins}
\usepackage{fontawesome}
\usepackage[misc]{ifsym}
\usepackage[capitalize]{cleveref}
\usepackage{balance}
\AtBeginDocument{%
  }

\copyrightyear{2025} 
\acmYear{2025} 
\setcopyright{cc}
\setcctype{by-nc}
\acmConference[CIKM '25]{Proceedings of the 34th ACM International Conference on Information and Knowledge Management}{November 10--14, 2025}{Seoul, Republic of Korea}
\acmBooktitle{Proceedings of the 34th ACM International Conference on Information and Knowledge Management (CIKM '25), November 10--14, 2025, Seoul, Republic of Korea}\acmDOI{10.1145/3746252.3761189}
\acmISBN{979-8-4007-2040-6/2025/11}

\newtcolorbox{mybox}[1][]{
breakable,
  arc=1mm,
  boxrule=1pt,
  colback=yellow!14,
  colframe=black!80,
  fonttitle=\bfseries,
  title=#1,
  left=1mm,
  right=1mm,
  top=1mm,
  bottom=1mm
}

\begin{document}

\title{Adapting Large Language Models to Log Analysis with Interpretable Domain Knowledge}

\author{Yuhe Ji}
\authornotemark[1]
\affiliation{%
  \institution{Nankai University}
  \country{Tianjin, China}
}
\email{jiyuhemail@foxmail.com}

\author{Yilun Liu}
\authornote{Equal contribution.}\authornote{Yilun Liu is the corresponding author.}\thanks{Nankai University is the first institution.}
\affiliation{%
  \institution{Nankai University}
  \country{Tianjin, China}
}
\affiliation{%
  \institution{Huawei}
  \country{Beijing, China}
}
\email{liuyilun3@huawei.com}

\author{Feiyu Yao}
\affiliation{%
  \institution{Huawei}
  \country{Beijing, China}
}
\email{frankyao.ece@gmail.com}

\author{Minggui He}
\author{Shimin Tao}
\affiliation{%
  \institution{Huawei}
  \country{Beijing, China}
}
\email{heminggui@huawei.com	}
\email{taoshimin@huawei.com}

\author{Xiaofeng Zhao}
\affiliation{%
  \institution{Huawei}
  \country{Beijing, China}
}
\email{zhaoxiaofeng14@huawei.com}

\author{Chang Su}
\affiliation{%
  \institution{Huawei}
  \country{Beijing, China}
}
\email{suchang8@huawei.com}

\author{Xinhua Yang}
\affiliation{%
  \institution{Huawei}
  \country{Beijing, China}
}
\email{yangxinhua2@huawei.com}

\author{Weibin Meng}

\affiliation{%
  \institution{Huawei}
  \country{Beijing, China}
}
\email{mengweibin3@huawei.com}

\author{Yuming Xie}
\affiliation{%
  \institution{Huawei}
  \country{Beijing, China}
}
\email{yuming.xie@huawei.com}

\author{Boxing Chen}
\affiliation{%
  \institution{Huawei Canada}
  \country{Montreal, Canada}
}
\email{boxing.chen@huawei.com}

\author{Shenglin Zhang}
\affiliation{%
  \institution{Nankai University}
  \country{Tianjin, China}
}
\email{zhangsl@nankai.edu.cn}

\author{Yongqian Sun}
\affiliation{%
  \institution{Nankai University}
  \country{Tianjin, China}
}
\email{sunyongqian@nankai.edu.cn}

\renewcommand{\shortauthors}{Ji, Liu, et al.}

\begin{abstract}
Log analysis represents a critical sub-domain within AI applications that facilitates automatic approaches to fault and error management of large-scaled software systems, saving labors of traditional manual methods. While existing solutions using large language models (LLMs) show promise, they are limited by a significant domain gap between natural and log languages (the latter contains rich domain-specific tokens such as status codes, IP addresses, resource pathes), which restricts their effectiveness in real-world applications. However, directly adapting general-purpose LLMs to log analysis using raw logs may degrade their performance due to inconsistent token distribution. In this paper, we present a domain adaptation approach that addresses these limitations by integrating interpretable domain knowledge into open-source LLMs through continual pre-training (CPT), which bridges this domain gap by adapting LLMs on interpretable natural texts with log knowledge (instead of raw logs) to reduce distribution discrepancy. To achieve this, we developed NLPLog, a comprehensive dataset containing over 250,000 question-answer pairs on log-related knowledge. Our resulting model, SuperLog, achieves the best performance across four log analysis tasks, with an average accuracy improvement of 12.01\% over the second-best model. Ablation study also suggests advantages of domain adaption using interpretable log knowledge over using raw logs. 
\end{abstract}



\maketitle

\section{Introduction}
Log analysis represents a critical sub-domain within AI applications, with significant implications for system reliability, security, and performance optimization. As computer systems and programs grow increasingly complex~\cite{jiang2024megascale,narayanan2021efficient,Jouppi2023TPUVA}, the inevitability of faults and errors necessitates innovative solutions that extend beyond the traditional reliance on experienced specialists sifting through extensive logs. This labor-intensive approach faces challenges due to the unpredictable nature of faults and errors, the sheer volume of logs, and the specialized knowledge required for effective log analysis.

In response to these challenges, there has been a burgeoning interest in leveraging large language models (LLMs) to enhance the efficiency and effectiveness of log analysis tasks. In this paper, LLMs are defined as language models with at least 7 billion (7B) parameters\cite{zhao2023survey}. Significant advancements have been made in several key log analysis tasks, including log parsing\cite{10172786,ma2024llmparser,liu2024logprompt}, log anomaly detection\cite{LogDAPT,10.1145/3588195.3595943,liu2024interpretable}, log classification, and log root cause analysis. Compared to smaller models, the advantages of LLMs in these tasks primarily lie in the interpretability of their analysis results\cite{liu2024interpretable} and their robust performance in online scenarios characterized by limited training data\cite{liu2024logprompt}. This shift towards LLM-based automated log analysis represents a significant trend in domain adaptation.

\begin{figure}[t!]
    \centering
  \includegraphics[width=\linewidth]{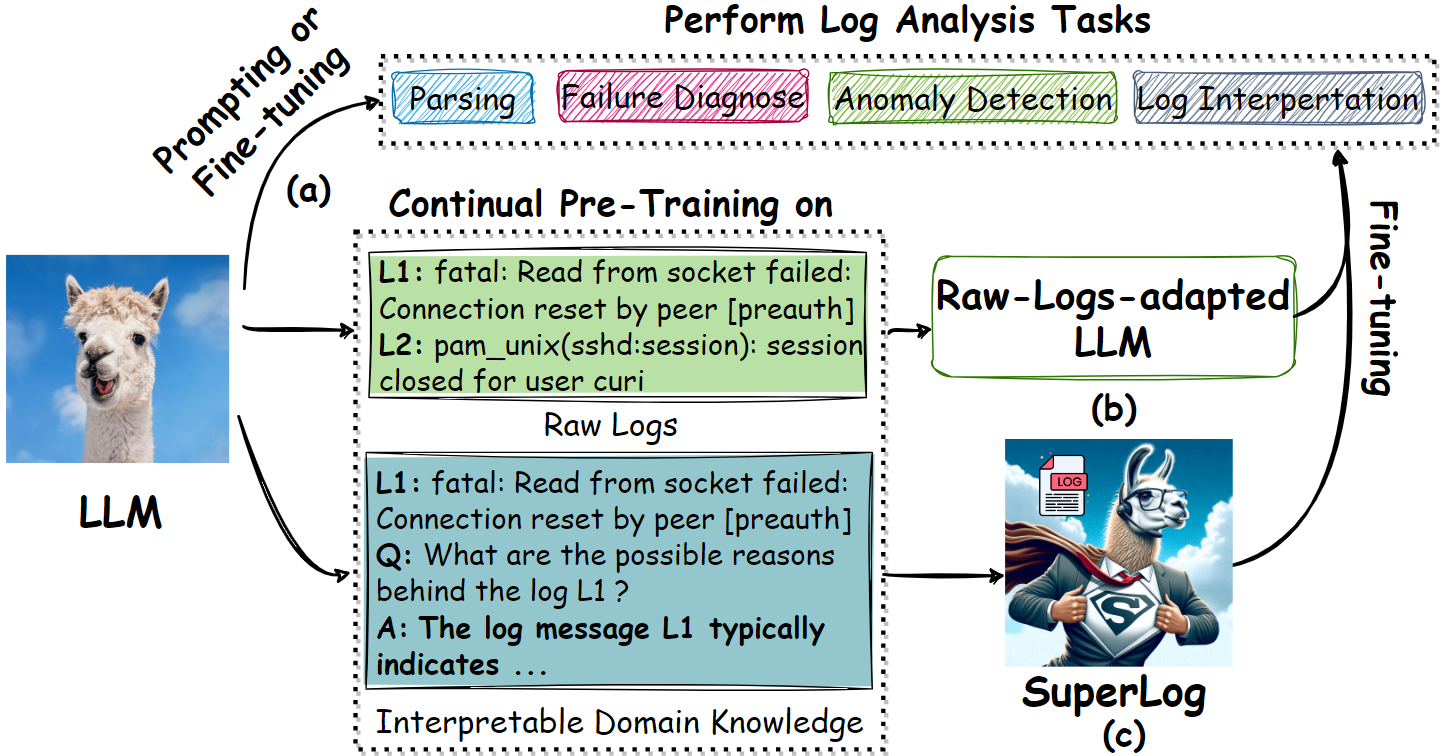}
  \caption{Illustration on differences of three LLM-based log analysis approaches: prompting or fine-tuning (a) on general-purpose LLMs, (b) on LLMs infusing raw logs and (c) on LLMs infusing interpretable domain knowledge (SuperLog).}
  \label{fig1}
\end{figure}

While these methods showcase promising advancements, their applicability in real-world scenarios remains constrained. As shown in Fig.~\ref{fig1}(a), most works attempt to directly prompt general-purpose LLMs to perform log tasks, which may lead to suboptimal performance due to the inherent gap between natural language and domain-specific language (i.e., logs). For instance, a study by~\cite{10.1145/3588195.3595943} illustrates that, requiring advanced LLMs to continuously summarize significant system events from historical logs and predict the current system state based on prompt skills, falls short of expectations. Similarly,~\cite{liu2024logprompt} attempts to equip propietaty LLMs with a set of advanced prompting strategies related to log tasks, achieving high performance in log parsing but still struggling with anomaly detection in zero-shot scenarios. This suboptimal performance may stem from a knowledge gap between logs and human language, as logs are typically concise, often grammatically incorrect, and lack comprehensive background information by their very nature~\cite{zhu2019tools, he2017drain, tao2022logstamp}. Powerful proprietary LLMs such as  may help mitigate this knowledge gap through their inference capabilities~\cite{qi2023loggpt, le2023log}. However, access to these proprietary LLMs is usually via APIs, necessitating an internet connection and retries upon access failures, which can hardly meet the security, robustness, and immediacy requirements of industrial applications. In contrast, open-source LLMs, such as the LLaMA model families~\cite{touvron2023llama}, offer greater deployment potential in real-world applications, yet the knowledge gap is even more pronounced for open-source LLMs attempting to perform log analysis tasks. This was noted by Liu \emph{et al.}~\cite{liu2024interpretable}, who utilized Vicuna~\cite{chiang2023vicuna} (fine-tuned from LLaMA) for log analysis and observed a marked performance gap when compared with a widely-used commercial language model available via API.

Before the emergence of LLMs, existing domain adaptation approaches primarily enhanced language models (ranging from approximately 0.5B to 1B parameters) through Continual Pre-Training (CPT)~\cite{gururangan2020don} directly on raw log data,  as decipted by the Raw-Logs-Adapted Logs in Fig.~\ref{fig1}(b). For example, Biglog~\cite{biglog} pre-trained the BERT model~\cite{devlin2018bert} on 83GB of raw log records collected from real-world devices~\cite{he2020loghub}. However, the limited interpretability of raw log data presents a significant challenge for language models, as their pre-trained corpora primarily consist of natural language texts. This discrepancy in the distribution of CPT datasets may lead to catastrophic forgetting~\cite{luo2023empirical}, a phenomenon where model performance deteriorates when newly added training data originate from a significantly different distribution. Furthermore, unlike BERT-like language models, LLMs are renowned for their ability to generate justifications alongside their predictions~\cite{liu2024interpretable}. The limited interpretability of domain knowledge during CPT may impede the interpretative capabilities of LLMs. Directly training on log data can reduce the likelihood of LLMs providing natural language explanations and justifications for their predictions, resulting in a notable decline in user-friendliness, as evidenced by our experimental results in Table~\ref{tab:abalation_exp}.

To address the challenge of insufficient domain knowledge in real-world log analysis using LLMs, this paper proposes a domain adaptation approach that enhances the performance of general-purpose open-source LLMs in log analysis tasks by integrating interpretable domain knowledge through CPT, as shown in Fig.~\ref{fig1}(c). Instead of trainging of raw logs, this approach creates an interpretable knowledge set for the log domain that can be effectively utilized to improve the performance of general-purpose LLMs. By incorporating this interpretable knowledge, we adapt LLMs to the target domain while preserving their inherent natural language comprehension and instruction-following abilities.
To facilitate reliable domain adaptation, we have developed a large-scale dataset called NLPLog, which contains over 250,000 question-and-answer pairs presented in natural language, emphasizing comprehension and analysis on real-world logs. This dataset serves as a valuable source of interpretable knowledge for domain adaptation of LLMs. As a result, our trained model, SuperLog, which undergoes the CPT phase using NLPLog, not only excels in executing log analysis tasks but also maintains a high degree of interpretability, aligning closely with industry demands for practical and understandable outcomes.
Our contributions are as follows:

\begin{itemize}
    \item We introduce a novel CPT paradigm that boosts large model performance by injecting interpretable knowledge. Ablation studies demonstrate that our new paradigm achieves a remarkable 23\% average performance improvement over traditional training methods.
    \item Building upon this paradigm, we developed SuperLog, which demonstrated superior few-shot and zero-shot performance across all four log analysis tasks. It surpassed the second-best model by an average of 12.01\% and showed exceptional performance on logs from unseen domains.
    \item We open-sourced a meticulously curated and large-scaled dataset, rich in log-related knowledge and derived from real-world log analysis practices, providing essential guidance for advancing new training paradigms\footnote{https://github.com/J-York/SuperLog}.
\end{itemize}
\section{RELATED WORK}

\subsection{LLMs \& Training Regimes}

LLMs have established themselves as pivotal tools in natural language processing, transforming our approach to language understanding and generation tasks. The training of LLMs typically involves multiple phases, each critical for achieving state-of-the-art performance.

The initial phase, known as pre-training, involves exposing the model to extensive amounts of unlabeled text data. This phase enables the model to learn general language patterns and representations, forming a robust linguistic foundation~\cite{zhang2022opt}. Pre-training is fundamental as it equips the model with the ability to understand and generate coherent text, which can be further refined for specific applications. To build the language contexts for LLMs over specialized domains, continual pre-training (CPT) is often employed. This technique involves updating the model's knowledge base with new domain-specific data, ensuring that the model adapts to the specialized language contexts~\cite{yildiz2024investigating}. CPT is especially crucial in fields with specialized language requirements that differ from general-purpose needs~\cite{biglog}.

Following pre-training and CPT, LLMs undergo a supervised fine-tuning (SFT) phase, where they are adapted to specific tasks using labeled datasets. This phase is crucial for task specialization, enabling the model to apply its broad linguistic knowledge to particular challenges such as sentiment analysis~, question answering, or text classification~\cite{sun2023text}. By fine-tuning on task-specific data, LLMs can achieve higher accuracy and versatility, making them feasible for a wide range of applications.

Our work redefines the paradigm of CPT for log analysis by infusing interpretable domain knowledge into LLMs. By constructing an interpretable dataset that combines log data with corresponding natural language explanations, the lack of log-related domain knowledge in general-purpose open-source LLMs is addressed.

\subsection{Log Analysis}
Log analysis encompasses log parsing, anomaly detection, fault diagnosis, and interpretation, ensuring efficient utilization of log data to enhance software system reliability and performance.

\subsubsection{Log Parsing}
Log parsing reduces log data to core elements by generating templates that capture essential patterns. Traditional methods, such as clustering~\cite{zhu2019tools}, heuristics~\cite{du2016spell}, and tree-structured approaches~\cite{he2017drain}, extract static components and replace variables with placeholders. Recent tools like LogParse~\cite{meng2020logparse} use word-level classifiers for dynamic pattern extraction. Advances by Huo~\emph{et al.}~\cite{huo2023semparser} and Li~\emph{et al.}~\cite{li2023did} focus on semantic modeling and classification. LLMs are increasingly applied, with techniques like LogPPT~\cite{le2023log} using prompt engineering, while Jiang~\emph{et al.}~\cite{jiang2024lilac} optimize parsing with adaptive mechanisms.

\subsubsection{Log-based Anomaly Detection}
Anomaly detection identifies irregular patterns in log data at session or template levels. Session-level methods classify entire sessions as anomalous if any template is unusual, using models like LSTM and CNNs (e.g., LogRobust~\cite{zhang2019robust}, DeepLog~\cite{du2017deeplog}). Innovations like Le~\emph{et al.}~\cite{le2021log} employ a BERT encoder to eliminate explicit parsing. Template-level methods, such as LogPrompt~\cite{liu2024interpretable} and RAGLog~\cite{pan2024raglog}, enhance detection with LLM-based prompting and retrieval-augmented generation.

\subsubsection{Log-based Fault Diagnosis}
Fault diagnosis identifies specific causes of anomalies, enabling timely issue resolution. Techniques involve root cause analysis through correlation and dependency mapping~\cite{sui2023logkg}. LLMs correlate error patterns with known fault signatures for precise diagnostics~\cite{liu2024loglm, cui2024logeval}, while machine learning tools offer predictive insights to reduce system downtime.

\subsubsection{Log Interpretation}
Log interpretation explains log events in natural language to enhance understanding. Advanced systems generate summaries, such as Liu~\emph{et al.}~\cite{liu2024logprompt}'s narrative descriptions. LLMs produce explanatory content for better insights~\cite{liu2024loglm}, and improved tools offer interactive systems for robust interpretation, boosting incident management and strategy formulation~\cite{chen2024automatic}.

\section{METHODOLOGY}

\begin{figure}[t!]
    \centering
  \includegraphics[width=\linewidth]{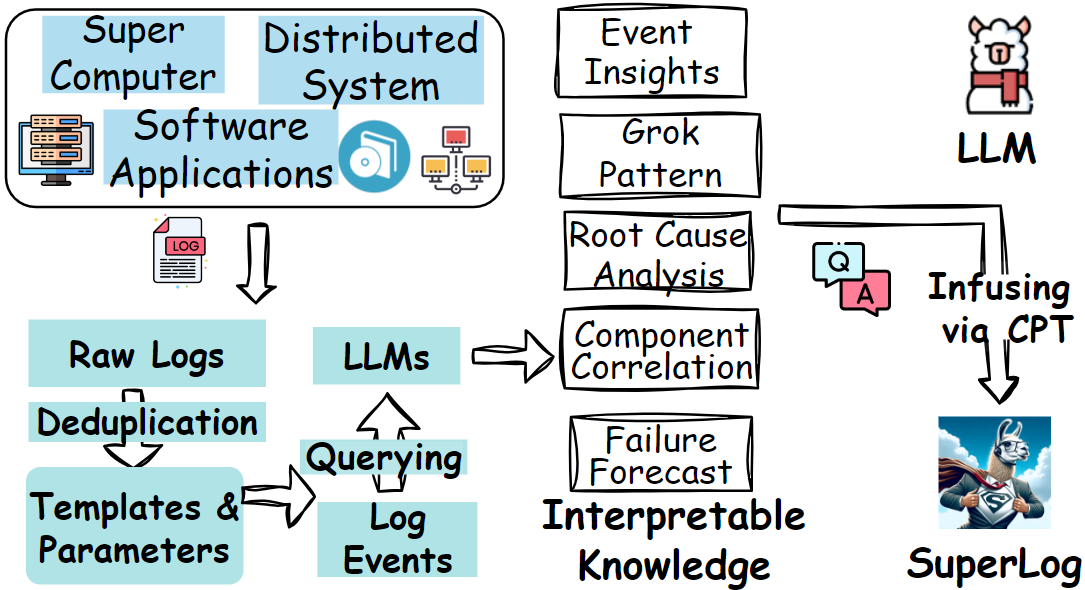}
  \caption{Illustration on the interpretable knowledge construction and continual pre-training of SuperLog.}
  \label{fig2}
\end{figure}

General-purpose LLMs often lack specialized domain knowledge, leading to suboptimal performance in log analysis~\cite{biglog}. To address this, we introduce SuperLog, an LLM enhanced with interpretable domain knowledge for log analysis. As shown in Fig.~\ref{fig2}, SuperLog undergoes a Continual Pre-Training (CPT) phase, where it acquires log-related knowledge while retaining its general language capabilities.

To enable this specialized training, we developed NLPLog, a large-scale dataset comprising natural language Q\&A pairs derived from real-world logs. This dataset serves as the foundation for imbuing LLMs with domain-specific knowledge, addressing the limitations of general-purpose models. Each training instance in NLPLog encompasses five key dimensions of log-related knowledge spanning 14 domains~\cite{he2020loghub}, ensuring the model's proficiency in addressing diverse queries. The answers provided in the dataset include comprehensive analysis, embedding interpretable knowledge directly into the training process and aligning the CPT phase with actual O\&M requirements.

Unlike traditional methods using raw log data, our approach employs natural language Q\&A pairs, improving interpretability and LLM compatibility. This bridges theory and practice, ensuring effective fine-tuning. The following section describes NLPLog's construction and the CPT process, detailing how they adapt general-purpose LLMs with domain knowledge.

\subsection{Construction of NLPLog}\label{sec:NLPLog}

 In this section, we introduce the construction process of NLPLog, the dataset for pre-training SuperLog. Particularly, we designed a meticulous framework to ensure data quality during the construction process.

To construct NLPLog dataset, we choose 14 different log domains from LogHub~\cite{he2020loghub}, an open-source dataset rich in real-world logs from different domains. These domains include operation systems, supercomputer, distributed system and software applications, thereby guaranteeing models trained on NLPLog dataset to focus on domain-invariant features and gain more robustness and generalization ability. However, since the log events are collected from real-world devices and systems within continuous time windows, there are large number of similar or even duplicated logs, which not only significantly increases the cost for creating NLPLog, but also may introduce unnecessary noises to the model during CPT phase. To address the aforementioned issues, we designed a data pre-processing framework which aims to select the most representative logs and generate interpretable knowledge from these logs by the form of Q\&A pairs, with three phases: Deduplication, Log Event Reconstruction, and Interpretable Knowledge Generation. Statistics of NLPLog is shown in Table~\ref{tab:NLPLog}.

\begin{table}[tp]
    \caption{Statistics of NLPLog, Our Constructed CPT Dataset}
    \centering
    \renewcommand{\arraystretch}{1.2} 
    \resizebox{0.9\linewidth}{!} {
        \begin{tabular} {p{0.25\linewidth}@{\hskip 0.01in}c@{\hskip 0.01in}c@{\hskip 0.01in}c}
        \toprule
        \multicolumn{1}{l}{\textbf{Domain}} & \multicolumn{1}{c}{\textbf{Log Count}} & \multicolumn{1}{c}{\textbf{Q\&A Pairs}} & \multicolumn{1}{c}{\textbf{Proportion}} \\
        \midrule
        \multicolumn{1}{l}{OpenSSH}      & \multicolumn{1}{c}{54}                   & \multicolumn{1}{c}{270}                & \multicolumn{1}{c}{0.19\%} \\
        \multicolumn{1}{l}{HDFS}         & \multicolumn{1}{c}{409}                  & \multicolumn{1}{c}{2,045}              & \multicolumn{1}{c}{1.54\%} \\ 
        \multicolumn{1}{l}{HPC}          & \multicolumn{1}{c}{159}                  & \multicolumn{1}{c}{795}                & \multicolumn{1}{c}{0.59\%} \\ 
        \multicolumn{1}{l}{Windows}      & \multicolumn{1}{c}{9,605}                & \multicolumn{1}{c}{48,025}             & \multicolumn{1}{c}{36.12\%} \\ 
        \multicolumn{1}{l}{Mac}          & \multicolumn{1}{c}{708}                  & \multicolumn{1}{c}{3,540}              & \multicolumn{1}{c}{2.63\%} \\ 
        \multicolumn{1}{l}{Thunderbird}  & \multicolumn{1}{c}{13,069}               & \multicolumn{1}{c}{65,345}             & \multicolumn{1}{c}{49.04\%} \\ 
        \multicolumn{1}{l}{Spark}        & \multicolumn{1}{c}{369}                  & \multicolumn{1}{c}{1,845}              & \multicolumn{1}{c}{1.38\%} \\ 
        \multicolumn{1}{l}{Linux}        & \multicolumn{1}{c}{654}                  & \multicolumn{1}{c}{3,270}              & \multicolumn{1}{c}{2.42\%} \\ 
        \multicolumn{1}{l}{Zookeeper}    & \multicolumn{1}{c}{104}                  & \multicolumn{1}{c}{520}                & \multicolumn{1}{c}{0.39\%} \\ 
        \multicolumn{1}{l}{HealthApp}    & \multicolumn{1}{c}{195}                  & \multicolumn{1}{c}{975}                & \multicolumn{1}{c}{0.73\%} \\ 
        \multicolumn{1}{l}{Hadoop}       & \multicolumn{1}{c}{270}                  & \multicolumn{1}{c}{1,350}              & \multicolumn{1}{c}{1.01\%} \\ 
        \multicolumn{1}{l}{BGL}          & \multicolumn{1}{c}{607}                  & \multicolumn{1}{c}{3,035}              & \multicolumn{1}{c}{2.26\%} \\ 
        \multicolumn{1}{l}{Android}      & \multicolumn{1}{c}{25,369}               & \multicolumn{1}{c}{126,845}            & \multicolumn{1}{c}{18.86\%} \\ 
        \multicolumn{1}{l}{Proxifier}    & \multicolumn{1}{c}{18}                   & \multicolumn{1}{c}{90}                 & \multicolumn{1}{c}{0.07\%} \\ 
        \bottomrule
        \end{tabular}
    }
    \renewcommand{\arraystretch}{1} 
    \label{tab:NLPLog}
\end{table}

\subsubsection{Deduplication}

Deduplication is an essential part of our framework, designed to minimize redundancy by identifying and extracting log templates from large quantities of semi-structured log data. Logs are composed of a fixed component (the template), originating from log statements that describe program execution events, and a dynamic component (the variable) that includes information such as LineID, Date, Time, and IP. Given that log templates provide essential insights into program execution and are significantly fewer in number compared to the total log entries, accurately extracting these templates improves the efficiency of log analysis by decreasing data volume and concentrating analysis on unique events.

For this purpose, we utilized LogPPT~\cite{10172786}, a sophisticated log template extraction algorithm. LogPPT leverages pre-trained language models and a small subset of labeled samples to identify log templates and the associated variables. This method enhances both the efficiency and accuracy of deduplication compared to traditional rule-based techniques. We used 2,000 manually parsed log entries from each domain available on LogHub as training data, and subsequently applied the trained model to the entire set of logs from these domains to derive their templates. After applying the log template extraction algorithm, we divided the logs into their template and variable components. Duplicate log templates were eliminated, resulting in 51,590 distinct log templates—a comprehensive collection of unique events that substantially reduces data redundancy and provides a robust foundation for further analysis.

\subsubsection{Log Event Reconstruction}

The Log Event Reconstruction process generates log events from a collection of log templates \(\{T_1, T_2, \dots, T_n\}\) and their associated variable groups. During template parsing, multiple log messages are parsed into a single template \(T_i\) and multiple variable groups \(\{G_1, G_2, \dots, G_p\}\), where each group \(G_j\) contains variables corresponding to the placeholders in \(T_i\). The process is as follows:

\textbf{Template Selection.} A log template \(T_i\) is selected from the collection.

\textbf{Variable Group Selection.} A variable group \(G_j\) is randomly selected from the set associated with \(T_i\). Each group \(G_j\) contains variables \(\{V_1, V_2, \dots, V_k\}\) matching the placeholders in \(T_i\).

\textbf{Placeholder Replacement.} The variables in \(G_j\) are used to replace the placeholders in \(T_i\), constructing a log event \(E\):

\[
E = \text{FixedPart}(T_i) + \{V_1, V_2, \dots, V_k\}.
\]

This ensures the generation of deduplicated, lossless log events, which serve as foundational data for training SuperLog.

\subsubsection{Interpretable Knowledge Generation}

To integrate interpretable and comprehensive log-related knowledge into the model for domain adaptation, we have distilled five key competency dimensions required by log analysis experts based on existing work. These dimensions are well-defined and reflect the core elements of current log analysis methodologies. We structure this knowledge as natural language Q\&A pairs, designing questions for each log and generating answers covering all five dimensions.

\textbf{Grok Pattern Parsing.} Using Grok~\cite{debnath2018loglens} is about deciphering the structure information of complex log data. It employs patterns to identify and extract details from log messages, making it easier to manage and analyze the data. This knowledge dimension focuses on identifying patterns within logs to simplify data extraction, making the log messages more manageable and facilitating efficient analysis.

\textbf{Log Event Insights.} 
Log Event Insights transform technical log data into clear, human-readable insights. By expanding on the semantic meaning of key log components, this dimension provides a more accessible understanding of log content, enhancing its relevance and clarity across diverse operational environments~\cite{10.1145/3377813.3381371}.

\textbf{Root Cause Analysis. }Root Cause Analysis is critical in log applications, as it identifies the underlying causes of system anomalies~\cite{suriadi2013root}. This knowledge dimension aids in pinpointing the source of issues, improving troubleshooting accuracy and enabling timely corrective actions.

\textbf{Component Correlation Analysis. }In complex systems, understanding the relationships between different components is vital~\cite{xie2021logm}. Component Correlation Analysis identifies and analyzes these interconnections within the logs, providing deeper insights into system interactions, which ultimately improves diagnostic precision and issue resolution.

\textbf{Potential Failure Forecast.} Failure Forecasting is critical in log analysis, involving the identification of patterns that precede potential failures~\cite{sipos2014log}. This knowledge dimension helps in predicting system failures by recognizing early warning signs in logs, allowing for proactive maintenance and preventing downtime.

Through learning from these five key dimensions, the model gains both structural and semantic understanding of real-world log events while developing its reasoning and associative capabilities for practical log analysis tasks.

To prevent overfitting caused by repetitive questioning, we developed 10 distinct question variations for each dimension. For every log data point, we randomly pick one variation and combine it with the log as an input prompt for a state-of-the-art LLM. The generated answers form the Q\&A pairs in our dataset, whose statistics are shown in Table~\ref{tab:NLPLog}. To ensure high quality, we include a human-calibration phase that filters out any responses that are inaccurate or stray from the intended context. This process guarantees that the resulting instruction pairs remain faithful to the original user posts and expert resolutions, preserving relevance and contextual integrity. An example entry from the dataset is shown below:
\begin{mybox}
\textbf{Question:} In your capacity as a performance tuning specialist in OpenSSH, what system performance anomalies could potentially be forecasted by this log?

\textbf{Log:} fatal: Read from socket failed: Connection reset by peer.

\textbf{Output:} The log message in OpenSSH indicates that the connection was unexpectedly terminated, which could suggest network or server issues.
\end{mybox}

\subsection{Continual Pre-Training}
\label{sec:CPT}

Continual Pre-Training (CPT) has emerged from recent advances in lifelong learning for NLP, strategically bridging the gap between general pre-training and task-specific fine-tuning~\cite{luo2023empirical, yildiz2024investigating}. This paradigm proves particularly critical for specialized domains like log analysis, where niche linguistic patterns (e.g., HEX strings, timestamps, severity codes) and structured semantics deviate significantly from general language distributions.

The self-supervised nature of CPT is achieved by training the LLaMA2 base model on NLPLog in a controlled environment, where the learning rate is set to 1e-5 and the training runs for 1.5 epochs. This controlled training ensures that the model does not overfit to any one specific task while still gaining substantial domain knowledge. The process enables the LLaMA2 model to learn both the syntax of logs and the specific knowledge contained within them, while retaining its general linguistic capabilities, thus enhancing its robustness.

\section{Experiments}

In this section, we assess the practical efficacy of SuperLog in log analysis tasks. The section is structured as follows: Section~\ref{sec:task_with_Superlog} demonstrates SuperLog's application in both zero-shot and few-shot learning contexts. Section~\ref{sec:details} provides implementation details, while Section~\ref{sec:rq_and_findings} outlines our research questions (RQs). Sections~\ref{sec:RQ1} through~\ref{sec:RQ3} present the experimental setup and findings corresponding to each RQ.

\subsection{Performing Log Analysis Tasks using SuperLog}\label{sec:task_with_Superlog}

To comprehensively evaluate SuperLog's capabilities, we employed few-shot learning for log parsing and anomaly detection, which benefit from specific in-domain sample to establish patterns, and zero-shot learning for fault diagnosis and interpretation, which leverage the model's pre-trained semantic understanding. This division reflects the nature of these tasks: parsing and anomaly detection require precise pattern recognition, while diagnosis and interpretation rely on contextual reasoning and generalization.

\subsubsection{Few-shot Learning Experiments} The first approach is few-shot Learning. Such a training approach fine-tune the model with a modest amount of in-domain task data, enabling SuperLog to swiftly apply the encapsulated log-related knowledge to these tasks.

For this purpose, we utilized popular public task-specific evaluation sets in log analysis. For log parsing task, we leveraged 2000 manually corrected parsing results provided by LogHub\_2k~\cite{he2020loghub} for each log domain and utilized the first 10\% logs to form instruction pairs for fine-tuning SuperLog. Instruction pairs for anomaly detection were derived from the BGL and Spirit benchmark datasets~\cite{oliner2007supercomputers}. Liu~\emph{et al.}\cite{liu2024interpretable} extracted log templates from these two datasets, respectively, releasing pairs of log templates and anomaly labels. We randomly selected approximately 10\% of each dataset to create instruction pairs, reserving the rest for evaluation. Each subset retained around 10\% abnormal samples, maintaining the original distribution of normal and anomalous logs. Using these datasets, SuperLog and other baseline models was fine-tuned over 3 epochs with a learning rate of 1e-5. This task-specific fine-tuning enabled the model to quickly adapt to the structured format and intricacies of each log domain, thereby enhancing its performance in downstream tasks.

\subsubsection{Zero-shot Learning Tasks}
The zero-shot learning approach is designed to enhance the model's capability to perform log fault diagnosis and log interpretation effectively, even without explicit task-specific training data. Instead of fine-tuning on particular log analysis tasks, the model is trained on a diverse set of open-domain instruction-following examples. This method aims to improve the model's versatility and its ability to analyze context in a new task instruction and give appropriate response fulfilling the requirement, particularly in scenarios where task-specific data is limited or unavailable.

To implement this approach, we utilized the AlpaCar\_1k dataset curated by Ge~\emph{et al.}~\cite{ge2024clustering}, which consists of 1,000 high-quality instruction-following examples. These instructions were meticulously selected by the authors to ensure both relevance and richness, forming a diverse and robust training set for our model. SuperLog and other baseline models was fine-tuned on this dataset over three epochs with a learning rate of 1e-5. This general-purpose instruction fine-tuning equips the model with the ability to follow a wide range of user instructions such as writing, math and common sense without log-related samples, making it highly interactive and adaptable. However, the model's zero-shot performance in log fault diagnosis and log interpretation relies heavily on the domain-specific knowledge embedded during the CPT phase, as this zero-shot training dataset does not directly incorporate task-specific data. Therefore, the effectiveness of zero-shot learning in these tasks hinges on the success of CPT in embedding comprehensive domain knowledge into the model.

\subsection{Implementation Details}\label{sec:details}

SuperLog utilizes the LLaMA-2-7B as its foundational model, which is a foundation LLM open-sourced by MetaAI~\cite{touvron2023llama2}. During the CPT phase, we employed the dataset shown in Table~\ref{tab:NLPLog}, setting the learning rate to 1e-5. The training was conducted for 1.5 epochs with a batch size of 16. During the instruction fine-tuning phase, we employed the experimental setup described in Section IV.A. Other parameters in both phases were kept at the default settings provided by LLaMA-Factory~\cite{zheng2024llamafactory}.

\subsection{Research Question}\label{sec:rq_and_findings}
In this section, we present the research questions(RQs) we addressed during the evaluation of SuperLog.

\textbf{RQ1:} Can SuperLog demonstrate strong performance on log-related downstream tasks?

\textbf{RQ2:} To what extent does training on a carefully constructed, interpretable dataset improve SuperLog's performance?

\textbf{RQ3:} How does SuperLog perform on logs from previously unseen domains?

\subsection{RQ1: Benchmarking on Log Analysis Capabilities}\label{sec:RQ1}
\leavevmode

\subsubsection{Few-shot Learning Performance}
\paragraph{Log Parsing} This benchmark assesses the performance of log parsing on the last 90\% of log entries from five distinct domains within the LogHub\_2k dataset. In this study, we evaluate SuperLog against 10 established log parsing approaches, which include cluster-based methods \cite{tang2011logsig,fu2009execution}, heuristic methods \cite{messaoudi2018search, du2016spell,makanju2009clustering}, tree-based methods \cite{zhang2017syslog,he2017drain}, machine learning methods \cite{meng2020logparse}, and LLM-based methods \cite{tao2022logstamp, liu2024logprompt}. Consistent with the experimental framework outlined by Liu \emph{et al.} \cite{liu2024interpretable}, all baseline models are trained using the initial 10\% of logs from each domain. An exception is LogPrompt \cite{liu2024logprompt}, which employs ChatGPT for log parsing without a training phase.

Based on the work of Liu \emph{et al.}~\cite{liu2024interpretable}, the evaluation criteria include both coarse-grained and fine-grained metrics. For the coarse-grained evaluation, the RandIndex~\cite{rand1971objective} is used. This metric evaluates the accuracy of log clustering by determining if logs with the same template are correctly grouped together, without considering the accuracy of the variables within the extracted templates. On the other hand, the fine-grained metric is the F1-score, which evaluates how accurately the variable parts in logs are identified. To compute the F1-score, the predicted log template is broken down into a sequence of tokens. For each token, the values $TP$, $TN$, $FP$, and $FN$ are counted. If a token is truly a variable and is correctly identified as such (or not), the value of $TP$ (or $FP$) is incremented by one. If a token is not a variable and is correctly predicted as not a variable (or incorrectly as a variable), the value of $TN$ (or $FN$) is incremented by one. The F1-score is calculated as the harmonic mean of Recall ($Recall = \frac{TP}{TP+FN}$) and Precision ($Precision = \frac{TP}{TP+FP}$).

SuperLog achieved outstanding results on the log parsing benchmark, surpassing all existing methods significantly in both coarse-level and fine-level evaluations. As shown in Table \ref{tab:logParsing_exp}, SuperLog outperformed the best baseline methods with an average improvement of 18.3\% in RandIndex (RI) and 13.3\% in F1-score. These superior results indicate that SuperLog is highly effective at accurately identifying variable components within logs and extracting precise coarse-level templates, setting a new standard in log parsing capabilities. 

\begin{table}[t!]
    \caption{Performance of Log Parsing under Few-shot Learning}
    \centering
    \resizebox{\linewidth}{!} {%
    \setlength{\tabcolsep}{2pt}
    \begin{tabular}{l@{\hskip 0.03in}c@{\hskip 0.03in}c@{\hskip 0.05in}|c@{\hskip 0.03in}c@{\hskip 0.05in}|c@{\hskip 0.03in}c@{\hskip 0.05in}|c@{\hskip 0.03in}c@{\hskip 0.05in}|c@{\hskip 0.03in}c@{\hskip 0.05in}c}
    \toprule
    \multirow{2}{*}{{\textbf{Methods}}} & \multicolumn{2}{>{\hspace{-0.5em}\centering}c}{\textbf{HDFS}} & \multicolumn{2}{>{\hspace{-0em}\centering}c}{\textbf{Hadoop}} & \multicolumn{2}{>{\hspace{-0.5em}\centering}c}{\textbf{Zookeeper}} & \multicolumn{2}{>{\hspace{-0.5em}\centering}c}{\textbf{Linux}} & \multicolumn{2}{>{\hspace{-0.5em}\centering}c}{\textbf{Proxifier}} \\ \cmidrule(l{0.3em}r{0.8em}){2-3} \cmidrule(l{0.3em}r{0.4em}){4-5} \cmidrule(l{0.3em}r{0.8em}){6-7} \cmidrule(l{0.3em}r{0.8em}){8-9} \cmidrule(l{0.3em}r{0.8em}){10-11}
     & \textbf{RI} & \textbf{F1} & \textbf{RI} & \textbf{F1} & \textbf{RI} & \textbf{F1} & \textbf{RI} & \textbf{F1} & \textbf{RI} & \textbf{F1} \\
    \midrule
    \addlinespace
    IPLoM & 0.914 & 0.389 & 0.636 & 0.068 & 0.787 & 0.225 & 0.695 & 0.225 & 0.822 & 0.500 \\
    LKE & 0.861 & 0.424 & 0.150 & 0.198 & 0.787 & 0.225 & 0.825 & 0.388 & 0.379 & 0.309 \\
    LogSig & 0.872 & 0.344 & 0.651 & 0.050 & 0.787 & 0.225 & 0.715 & 0.146 & 0.559 & 0.339 \\
    FT-tree & 0.908 & 0.385 & 0.668 & 0.046 & 0.773 & 0.186 & 0.709 & 0.211 & 0.722 & 0.420 \\
    Spell & 0.871 & 0.000 & 0.721 & 0.058 & 0.102 & 0.045 & 0.706 & 0.091 & 0.621 & 0.000 \\
    Drain & 0.914 & 0.389 & 0.647 & 0.068 & 0.787 & 0.225 & 0.695 & 0.225 & 0.822 & 0.500 \\
    MoLFI & 0.871 & 0.000 & 0.699 & 0.095 & 0.899 & 0.000 & 0.410 & 0.026 & 0.621 & 0.000 \\
    LogParse & 0.907 & 0.632 & 0.349 & 0.502 & 0.982 & 0.348 & 0.825 & 0.588 & 0.490 & 0.334 \\
    LogStamp & 0.954 & 0.523 & 0.927 & 0.594 & 0.992 & 0.275 & 0.760 & 0.658 & 0.811 & 0.438 \\
    LogPrompt & 0.890 & 0.863 & 0.879 & 0.763 & 0.948 & \textbf{0.889} & 0.758 & 0.766 & 0.567 & 0.653 \\
    \hdashline
    \noalign{\vskip 2pt}
    \textbf{SuperLog} & \textbf{0.979} & \textbf{0.988} & \textbf{0.982} & \textbf{0.942} & \textbf{0.998} & 0.815 & \textbf{1.000} & \textbf{0.914} & \textbf{0.998} & \textbf{0.939} \\
    \bottomrule
    \multicolumn{11}{l}{$^{\mathrm{a}}$ \textbf{RI} stands for coarse-level RandIndex. \textbf{F1} stands for fine-level F1-score.} \\
    \end{tabular}
    }
    \label{tab:logParsing_exp}
\end{table}

\paragraph{Log Anomaly Detection} This evaluation compares SuperLog with both template-level methods~\cite{liu2024interpretable} and session-level methods~\cite{du2017deeplog, meng2019loganomaly, zhang2019robust}. Accordingly, the evaluation is divided into two parts: template-level and session-level. 

For the template-level evaluation, the test set consists of the split template-label pairs, representing approximately 90\% of the templates extracted by Liu~\emph{et al.}~\cite{liu2024interpretable} from the BGL and Spirit datasets.

For session-level evaluation, log sessions were grouped into fixed windows of 100 logs from BGL and Spirit. The first 4000 logs were used for training, while the remaining logs formed the test set. Training logs were excluded to prevent data leakage, yielding 40,521 test sessions for BGL and 7,515 for Spirit. For both template-level and session-level assessments, we employ the F1-score of anomalies as the evaluation metric, as detailed in the previous section. 

The evaluation result is show in Table~\ref{tab:anomaly_exp}. From an overall perspective, selecting only a small subset of logs in sequence as the training set presents a significant challenge for most log anomaly detection methods. The sequential selection, as opposed to random selection, restricts the model to learning from a short segment of the log sequence, making it difficult to capture the overall distribution patterns of the logs. However, through the injection of interpretable knowledge, SuperLog demonstrates a strong understanding of log data, enabling it to extrapolate learning results from limited data. Ultimately, SuperLog outperforms existing state-of-the-art algorithms across all evaluation metrics, with particularly significant improvements observed on large-scale log datasets, such as the Spirit dataset.

\begin{table}[t!]
    \caption{Performance of Anomaly Detection under Few-shot Learning}
    \centering
    \resizebox{0.84\linewidth}{!} {%
    \begin{tabular}{l@{\hskip 0.15in}c@{\hskip 0.1in}c@{\hskip 0.1in}c@{\hskip 0.15in}c@{\hskip 0.1in}c@{\hskip 0.1in}c}
    \toprule
    \multirow{2}{*}{\textbf{Model}} & \multicolumn{3}{c}{\hspace{-1.5em}\textbf{BGL}}       & \multicolumn{3}{c}{\hspace{-0.2em}\textbf{Spirit}} \\ \cmidrule(l{0em}r{1.5em}){2-4} \cmidrule(l{-0.5em}r{0.2em}){5-7} 
                                   & \hspace{0.6em}\textbf{S-F1}$^{\mathrm{a}}$ & \textbf{T-F1} &   & \textbf{S-F1} & \textbf{T-F1} &    \\ \midrule
    LogBERT~\cite{guo2021logbert}               & 0.108             & -         &    & 0.049             & -         &    \\
    LogAnomaly~\cite{meng2019loganomaly}        & 0.129             & -        &    & 0.138             & -         &    \\
    LogRobust~\cite{zhang2019robust}            & 0.077             & -         &    & 0.045             & -         &    \\
    LogPrompt~\cite{liu2024logprompt}            & 0.129             & 0.067          &    & 0.122             & 0.050          &    \\
    \hdashline
    \noalign{\vskip 2pt}
    \textbf{SuperLog}        & \textbf{0.147}    & \textbf{0.262}          &    & \textbf{0.333}    & \textbf{0.300}          &    \\ \bottomrule 

    \multicolumn{7}{l}{$^{\mathrm{a}}$ \textbf{S-F1}/\textbf{T-F1} means F1-Score in session/template-level.} \\
    \end{tabular}
    }
    \label{tab:anomaly_exp}
\end{table}

\subsubsection{Zero-shot Learning Performance}
\leavevmode

\paragraph{Log Interpretation} Log interpretation and understanding are vital for extracting meaningful insights from log data. Drawing on Liu's research~\cite{liu2024loglm}, we define the log interpretation capabilities of language models in two key aspects: usefulness, where the model's interpretation should encompass domain understanding, extract key information, and assist analysts; and readability, where the output should be concise, clear, and expressed in natural language, avoiding confusion. To evaluate these capabilities, we selected a dataset of 100 log entries and tasked SuperLog with explaining the events each log represents. Four experienced log maintenance experts assessed the model's outputs comprehensively, using predefined criteria. The evaluation focused on usefulness and readability, with scores ranging from 1 to 5. Finally, we calculated the average score across all 100 responses to measure the model's overall performance.

We selected Qwen2-0.5B, Qwen2-1.5B~\cite{yang2024qwen2}, LLaMA3.1-8B, and OWL-7B~\cite{guoowl} as baseline models for comparison. Qwen2 is a general-purpose LLM family open-sourced by Alibaba, demonstrating strong performance across various domains. OWL-7B, on the other hand, is a domain-specific LLM designed for Q\&A in IT operations. As shown in Table~\ref{tab:logIntrepre_exp}, the experimental results reveal two key findings. First, SuperLog's readability not only exceeds all baselines but also surpasses industrial-grade LLaMA3.1-8B by 9.1\%. Second, our model achieves 24.5\% higher usefulness score than the best-performing general model Qwen2-0.5B, and outperforms the domain-specialized OWL-7B by 9.8\% in this critical metric. Such leap-forward performance stems from our dual-phase optimization: the CPT phase injects domain interpretability through curated knowledge distillation, while the subsequent SFT ensures linguistic fluency via high-quality instruction tuning.

\begin{table}[t]
\caption{Performance of Log Interpretation under Zero-shot Learning}
\centering
\resizebox{0.70\linewidth}{!}{
    \begin{tabular}{@{}lcc@{}}
    \toprule
    \textbf{Models}          & \textbf{Usefulness} & \textbf{Readability} \\ \midrule
    Qwen2-0.5B    & 3.845      & 4.125       \\
    \noalign{\vskip 1pt}
    Qwen2-1.5B    & 3.510      & 4.200       \\
    \noalign{\vskip 1pt}
    LLaMA3.1-8B   & 3.830      & 4.380       \\
    OWL-7B            & 4.034      & 3.950                \\
    \hdashline
    \noalign{\vskip 2pt}
    SuperLog(Ours) & \textbf{4.430}     & \textbf{4.780}     \\
    \bottomrule
    \end{tabular}
}
\label{tab:logIntrepre_exp}
\end{table}

\paragraph{Log-based Failure Diagnosis} In this section, our experimental setup for log-based failure diagnosis aligns with the LogEval benchmark~\cite{cui2024logeval}, a comprehensive suite designed to evaluate the capabilities of LLMs in log analysis tasks. We utilize log datasets sourced from Alibaba Cloud and China Mobile~\cite{cui2024logeval}, which capture diverse real-world scenarios. From these datasets, we selected 4,000 representative failure logs to construct the test set, enabling a robust assessment of the model's performance.

For our baseline, we selected open-source LLMs, including general-purpose models such as BaiChuan2-13b~\cite{yang2023baichuan}, AquilaChat-7b~\cite{baai2023aquilachat}, LLaMa2-70b~\cite{touvron2023llama2}, Qwen1.5-72b~\cite{bai2023qwen}, InternLM2-7b, InternLM2-20b~\cite{cai2024internlm2}, and Mistral-7b~\cite{jiang2023mistral}, as well as DeVops-7b and DeVops-14b~\cite{ebert2016devops}, which are specifically trained for O\&M tasks.

\begin{table}[tbp]
    \caption{Performance of Failure Detection and Failure Diagnosis in F1 Score}
    \centering
    \resizebox{0.75\linewidth}{!} {%
    \begin{tabular}{lc@{\hskip 0.1in}c@{\hskip 0.1in}c@{\hskip 0.1in}c}
    \toprule
    
    \multirow{1}{*}{\textbf{Methods}} & 
    \multicolumn{1}{c}{\begin{tabular}[c]{@{}c@{}}  \textbf{Detection}\\  \end{tabular} }  & 
    \multicolumn{1}{c}{\begin{tabular}[c]{@{}c@{}} \hspace{0.1em} \textbf{Diagnosis}\\ \end{tabular} }  \\

    \midrule
    BaiChuan2-13B           & 0.0            & 0.0   \\
    DeVops-7B               & 0.037          & 0.357 \\
    AquilaChat-7B           & 0.042          & 0.348 \\
    LLaMa2-70B              & 0.044          & 0.291 \\
    DeVops-14B              & 0.055          & 0.416 \\
    Qwen1.5-72B             & 0.063          & 0.423 \\
    InternLM2-7B            & 0.075          & 0.284 \\
    InternLM2-20B           & 0.089          & 0.425 \\
    Mistral-7B              & 0.092          & 0.284 \\
    \hdashline
    
    \noalign{\vskip 2pt}
    \textbf{SuperLog (ours)} & \textbf{0.117} & \textbf{0.500} \\
        \bottomrule
    \end{tabular}
    }
    \label{tab:anomaly_diagnose}
\end{table}

The experimental setup includes two stages: first, a binary classification task to determine whether a log entry represents a failure (failure detection), followed by a multi-class classification task to identify the specific type of fault (fault diagnosis). The final experimental results are shown in Table~\ref{tab:anomaly_diagnose}. SuperLog outperformed all baseline algorithms in both log failure detection and log-based failure diagnosis. Compared with other LLMs, SuperLog demonstrated superior performance. Similarly, when compared with models specifically trained for O\&M tasks, SuperLog also achieved better results. These findings validate the advanced nature of the NLPLog dataset we developed and highlight the importance of injecting interpretable knowledge to enable large models to efficiently adapt to domain-specific tasks.

\subsection{RQ2: Ablation Study on Training Datasets and Methods}\label{sec:RQ2}

\begin{table}[t!]
\caption{Ablation Study of SuperLog: Eliminating Interpretable Knowledge or CPT Phase}
\centering
\resizebox{0.76\linewidth}{!} {%
\begin{tabular}{@{}lcccc@{}}
\toprule
\textbf{Methods}  & \textbf{Parsing}        & \textbf{AD}  & \textbf{FD}             & \textbf{Inter}          \\ \midrule
SuperLog & \textbf{0.920} & \textbf{0.117} & \textbf{0.500} & \textbf{3.895} \\

w/o IK   & 0.906          & 0.096          & 0.382          & 3.054           \\

w/o CPT  & 0.881          & 0.090          & 0.311          & 3.273          \\ \bottomrule
\end{tabular}
}
\begin{flushleft}
\textbf{Parsing}: Log Parsing; \textbf{AD}: Anomaly Detection; \textbf{FD}: Log-based Failure Diagnose; \textbf{Inter}: Log Interpretation; \textbf{w/o IK}: pre-training only on logs in NLPLog; \textbf{w/o CPT}: no continual pre-training phase.
\end{flushleft}
\label{tab:abalation_exp}
\end{table}

\subsubsection{Evaluation Setting}
To thoroughly assess SuperLog's performance, we conducted two ablation experiments. 
(1) \textbf{SuperLog w/o CPT}: In this setup, we fine-tuned the LLaMA2-7B model on the Alpaca-1k dataset to instill instruction-following capabilities, omitting the continual pre-training (CPT) phase. 
(2) \textbf{SuperLog w/o IK}: Here, we retain LLaMA2-7B as the base model but omit the extra interpretable knowledge that was previously distilled from a proprietary LLM. Instead, we feed the CPT stage with only the deduplicated raw logs sourced directly from LogHub.
Consistent with prior sections, we evaluated both variants across four tasks: log parsing, log anomaly detection, log-based fault diagnosis, and log interpretation.

\subsubsection{Result} The experimental results are presented in Table~\ref{tab:abalation_exp}. We evaluate the performance of SuperLog across various tasks, including log parsing, log anomaly detection, and log-based fault diagnosis, using the F1-score as the metric. For log interpretation tasks, we assess the model based on the average of usefulness and readability. 

SuperLog outperformed other models in all tasks. When compared to models without the CPT phase, SuperLog exhibited superior performance, as it effectively acquired more domain-specific information during training, transitioning from a general model to a domain-specific one. In contrast to the dataset where only raw log data was used for CPT, SuperLog's performance was enhanced by the incorporation of interpretable knowledge during the CPT phase. Furthermore, models using CPT with raw log texts showed improvement in the three log analysis tasks. However, their performance in log interpretation was lower than that of models without the CPT phase. This suggests that while CPT can support knowledge injection, it may also lead to catastrophic forgetting. NLPLog addresses this by constructing Q\&A pairs, bridging the gap between domain-specific knowledge and natural language expressions, thus facilitating interpretable domain knowledge injection during the CPT phase. The results of the ablation study confirm the effectiveness of this new paradigm for domain knowledge injection, demonstrating that the integration of interpretable knowledge significantly enhances the model's specialized capabilities in the target domain.

\subsection{RQ3: Benchmarking on unseen Domain Logs}\label{sec:RQ3}

\subsubsection{Evaluation Setting} In this section, we evaluate the performance of SuperLog on unseen log domains by conducting experiments on two datasets—Apache and OpenStack—that were not included in NLPLog. Since these datasets do not have ground truth labels, we compare the model's output with results generated by an advanced LLM to serve as a reference for evaluation. Specifically, we replicate the log parsing experiment setup used in previous studies, where results generated by an advanced LLM for Apache and OpenStack logs are treated as the target labels. Different large models are then applied to perform log parsing tasks on these datasets, and their outputs are compared against the advanced LLM-generated references. To quantify the similarity between the model outputs and the references, we compute ROUGE scores, with ROUGE-1 measuring unigram overlap and ROUGE-L assessing the longest common subsequences. These metrics provide a quantitative evaluation of the quality of machine-generated text in the absence of human-labeled references.

\begin{table}[t!]
    \caption{Evaluation Of SuperLog on Unseen Domains}
    \centering
    \resizebox{0.9\linewidth}{!} {%
    \begin{tabular}{l@{\hskip 0.1in}c@{\hskip 0.05in}c@{\hskip 0.05in}c@{\hskip 0.1in}c@{\hskip 0.05in}c@{\hskip 0.05in}c}
    \toprule
    \multirow{2}{*}{\textbf{Methods}} & \multicolumn{3}{c}{\hspace{-1.5em}\textbf{Apache}}       & \multicolumn{3}{c}{\hspace{-0.2em}\textbf{OpenStack}} \\ \cmidrule(l{0em}r{1.5em}){2-4} \cmidrule(l{-0.5em}r{0.2em}){5-7} 
                                   & \hspace{0.6em}\textbf{Rouge-1} & \textbf{Rouge-L} &   & \textbf{Rouge-1} & \textbf{Rouge-L} &    \\ \midrule
    LLaMa3.1-8B               & 35.534             & 12.314         &    & 32.015             & 11.395         &    \\
    Qwen2-0.5B        & 32.686             & 11.917        &    & 34.456             & 14.665         &    \\
    Qwen2-1.5B            & 41.507             & 16.147         &    & 40.540             & 16.013         &    \\
    OWL-7B            & 48.763             & 30.841          &    & 44.819             & 23.832          &    \\
    \hdashline
    \noalign{\vskip 2pt}
    \textbf{SuperLog}          & \textbf{51.703}             & \textbf{42.224}          &    & \textbf{52.348}             & \textbf{34.071}          &     \\ \bottomrule 

    \end{tabular}
    }
    \label{tab:unseen_domain}
\end{table}

\subsubsection{Results} The performance of SuperLog on unseen domains is shown in Table~\ref{tab:unseen_domain}. SuperLog's ROUGE scores are consistently higher than those of existing baseline algorithms, with an improvement of approximately 22.4\% over the second-best performing model, OWL, significantly outperforming the LLaMA 3.1 and Qwen 2 series models. The experiment demonstrates that SuperLog possesses exceptional log understanding capabilities, performing well even on unseen domains. Its outputs are highly aligned with human tendencies and show strong consistency with professional annotations from operations engineers. This indicates that SuperLog is not only capable of achieving excellent performance in familiar domains but is also effective in understanding and processing log data in unseen domains.

\section{Discussion}
\subsection{Implications of Findings} 
\subsubsection{Effective Domain Adaptation Through Interpretable Knowledge} Our approach demonstrates the critical importance of innovative domain adaptation strategies that bridge the significant gap between natural language and logs. SuperLog's success stems from its ability to retain the robust natural language comprehension inherent in general-purpose LLMs while simultaneously acquiring specialized proficiency in log analysis. This balance is achieved through our domain adaptation approach using continual pre-training (CPT) with the NLPLog dataset, where domain knowledge is distilled into interpretable question-and-answer pairs expressed in natural language rather than training directly on raw logs.
By adapting the linguistic structure of the base model (LLaMA2-7B) with structured log-related insights presented in natural language format, SuperLog effectively reduces distribution discrepancy and avoids the pitfalls of catastrophic forgetting—a common challenge when adapting LLMs to domains with significantly different data distributions. Our experimental results, showing an average improvement of 12.01\% over existing methods, validate that domain adaptation using interpretable knowledge can significantly outperform traditional approaches that rely on raw logs. This finding suggests that effective domain adaptation need not come at the expense of a model's general capabilities. Instead, it can serve as a complementary layer that enriches the model's versatility while bridging domain gaps.
\subsubsection{Interpretability and Transparency} A hallmark of our domain adaptation approach is its emphasis on interpretability, achieved by embedding domain knowledge in a form that aligns with human reasoning processes. Unlike traditional domain adaptation approaches that rely on raw log data, our method ensures that SuperLog not only excels in log analysis tasks but also delivers outcomes that are transparent and justifiable. This addresses a fundamental limitation of previous domain adaptation methods that sacrifice interpretability when working with domain-specific data like logs.
This interpretability advantage is particularly evident in tasks like log interpretation and fault diagnosis, where the model provides natural language explanations alongside its predictions, as demonstrated in our zero-shot learning experiments (Section~\ref{sec:RQ1}). SuperLog's ability to articulate its reasoning bridges the gap between complex log data and actionable insights, empowering engineers and analysts to validate and act upon its outputs with confidence. Beyond log analysis, this finding has broader implications for domain adaptation approaches across specialized fields, particularly in safety-critical systems, such as autonomous vehicles or cybersecurity, where opaque "black-box" models are often met with skepticism. By prioritizing interpretability in domain adaptation, our approach sets a precedent for designing AI systems that meet both performance and accountability standards, potentially influencing regulatory frameworks and industry best practices.
\subsection{Threats to Validity}
Despite the promising results achieved by SuperLog, several limitations need to be acknowledged, which could guide future research directions.
\subsubsection{Generalizability} Although SuperLog performed well on unseen domains, its performance might degrade with logs that have significantly different characteristics or structures. In Section~\ref{sec:RQ3}, we treat the target log domain as label-free and adopt reference solutions produced by a state-of-the-art proprietary language system; ROUGE-1 and ROUGE-L are used to quantify alignment with these references. Nevertheless, high surface similarity to these proxy answers does not guarantee the method has adequately grasped the underlying task. We therefore plan to expand evaluation to additional domains and employ complementary metrics before drawing conclusions about generalizability.

\subsubsection{Hallucination} The phenomenon of hallucination in LLMs presents a significant limitation, particularly in applications requiring high accuracy and reliability, such as log-based fault diagnosis. Hallucination refers to the model's tendency to generate content that is coherent but factually incorrect or inconsistent with the provided source content~\cite{zhang2023siren}. In this case, the model may generate responses that are difficult to directly assess for correctness, potentially affecting the judgment of the operations staff.

\section{Conclusion}
In this paper, we present a novel domain adaptation approach to log analysis that effectively bridges the significant gap between natural language and log languages. Our method enhances the capabilities of LLMs by incorporating interpretable domain knowledge through continual pre-training (CPT), rather than directly adapting models on raw logs. This innovative domain adaptation strategy reduces distribution discrepancy by training LLMs on interpretable natural texts containing log knowledge, thereby preserving their natural language capabilities while acquiring domain expertise.
A key element of our approach is the development of the NLPLog dataset, which contains over 250,000 question-answer pairs, offering a rich repository of log-related knowledge presented in natural language format. By utilizing this domain adaptation paradigm and the NLPLog dataset, we trained SuperLog, an LLM specifically designed for log analysis tasks. Our experimental results demonstrate that SuperLog outperforms existing state-of-the-art methods across four log analysis tasks, achieving an average accuracy improvement of 12.01\% over the second-best model, including robust performance on logs from previously unseen domains.
These results confirm that our domain adaptation approach successfully infuses domain-specific knowledge while maintaining the interpretability advantages of LLMs. The ablation studies further validate the superiority of adapting models using interpretable log knowledge compared to traditional approaches using raw logs. To encourage further research in domain adaptation techniques and log analysis, we have made the NLPLog dataset publicly available for training large models on domain-specific tasks.

\section*{Disclosure}
Generative AI was used for polishing writing and assisting in data generation.

\bibliographystyle{ACM-Reference-Format}
\balance
\bibliography{sample-base,software}


\begin{thebibliography}{64}


\ifx \showCODEN    \undefined \def \showCODEN     #1{\unskip}     \fi
\ifx \showISBNx    \undefined \def \showISBNx     #1{\unskip}     \fi
\ifx \showISBNxiii \undefined \def \showISBNxiii  #1{\unskip}     \fi
\ifx \showISSN     \undefined \def \showISSN      #1{\unskip}     \fi
\ifx \showLCCN     \undefined \def \showLCCN      #1{\unskip}     \fi
\ifx \shownote     \undefined \def \shownote      #1{#1}          \fi
\ifx \showarticletitle \undefined \def \showarticletitle #1{#1}   \fi
\ifx \showURL      \undefined \def \showURL       {\relax}        \fi
\providecommand\bibfield[2]{#2}
\providecommand\bibinfo[2]{#2}
\providecommand\natexlab[1]{#1}
\providecommand\showeprint[2][]{arXiv:#2}

\bibitem[Bai et~al\mbox{.}(2023)]%
        {bai2023qwen}
\bibfield{author}{\bibinfo{person}{J. Bai}, \bibinfo{person}{S. Bai}, \bibinfo{person}{Y. Chu}, {et~al\mbox{.}}} \bibinfo{year}{2023}\natexlab{}.
\newblock \showarticletitle{Qwen technical report}.
\newblock \bibinfo{journal}{\emph{arXiv preprint arXiv:2309.16609}} (\bibinfo{year}{2023}).
\newblock


\bibitem[{Beijing Academy of Artificial Intelligence}(2023)]%
        {baai2023aquilachat}
\bibfield{author}{\bibinfo{person}{{Beijing Academy of Artificial Intelligence}}.} \bibinfo{year}{2023}\natexlab{}.
\newblock \bibinfo{title}{{Aquilachat}}.
\newblock
\urldef\tempurl%
\url{https://model.baai.ac.cn/model-detail/100101}
\showURL{%
\tempurl}


\bibitem[Cai et~al\mbox{.}(2024)]%
        {cai2024internlm2}
\bibfield{author}{\bibinfo{person}{Z. Cai}, \bibinfo{person}{M. Cao}, \bibinfo{person}{H. Chen}, {et~al\mbox{.}}} \bibinfo{year}{2024}\natexlab{}.
\newblock \showarticletitle{Internlm2 technical report}.
\newblock \bibinfo{journal}{\emph{arXiv preprint arXiv:2403.17297}} (\bibinfo{year}{2024}).
\newblock


\bibitem[Chen et~al\mbox{.}(2024)]%
        {chen2024automatic}
\bibfield{author}{\bibinfo{person}{Y. Chen}, \bibinfo{person}{H. Xie}, \bibinfo{person}{M. Ma}, {et~al\mbox{.}}} \bibinfo{year}{2024}\natexlab{}.
\newblock \showarticletitle{Automatic Root Cause Analysis via Large Language Models for Cloud Incidents}. In \bibinfo{booktitle}{\emph{Proc. of the European Conference on Computer Systems}}.
\newblock


\bibitem[Chiang et~al\mbox{.}(2023)]%
        {chiang2023vicuna}
\bibfield{author}{\bibinfo{person}{W.L. Chiang}, \bibinfo{person}{Z. Li}, \bibinfo{person}{Z. Lin}, {et~al\mbox{.}}} \bibinfo{year}{2023}\natexlab{}.
\newblock \showarticletitle{Vicuna: An Open-Source Chatbot Impressing GPT-4 with 90\%* ChatGPT Quality}.
\newblock \bibinfo{journal}{\emph{See https://vicuna. lmsys. org}} (\bibinfo{year}{2023}).
\newblock


\bibitem[Cui et~al\mbox{.}(2024)]%
        {cui2024logeval}
\bibfield{author}{\bibinfo{person}{T. Cui}, \bibinfo{person}{S. Ma}, \bibinfo{person}{Z. Chen}, {et~al\mbox{.}}} \bibinfo{year}{2024}\natexlab{}.
\newblock \showarticletitle{LogEval: A Comprehensive Benchmark Suite for Large Language Models In Log Analysis}.
\newblock \bibinfo{journal}{\emph{arXiv preprint arXiv:2407.01896}} (\bibinfo{year}{2024}).
\newblock


\bibitem[Debnath et~al\mbox{.}(2018)]%
        {debnath2018loglens}
\bibfield{author}{\bibinfo{person}{B. Debnath}, \bibinfo{person}{M. Solaimani}, \bibinfo{person}{M.A.G. Gulzar}, {et~al\mbox{.}}} \bibinfo{year}{2018}\natexlab{}.
\newblock \showarticletitle{LogLens: A Real-Time Log Analysis System}. In \bibinfo{booktitle}{\emph{Proc. of the IEEE International Conference on Distributed Computing Systems (ICDCS)}}.
\newblock


\bibitem[Devlin(2018)]%
        {devlin2018bert}
\bibfield{author}{\bibinfo{person}{J. Devlin}.} \bibinfo{year}{2018}\natexlab{}.
\newblock \showarticletitle{Bert: Pre-training of Deep Bidirectional Transformers for Language Understanding}.
\newblock \bibinfo{journal}{\emph{arXiv preprint arXiv:1810.04805}} (\bibinfo{year}{2018}).
\newblock


\bibitem[Du and Li(2016)]%
        {du2016spell}
\bibfield{author}{\bibinfo{person}{M. Du} {and} \bibinfo{person}{F. Li}.} \bibinfo{year}{2016}\natexlab{}.
\newblock \showarticletitle{Spell: Streaming Parsing of System Event Logs}. In \bibinfo{booktitle}{\emph{Proc. of the IEEE International Conference on Data Mining (ICDM)}}.
\newblock


\bibitem[Du et~al\mbox{.}(2017)]%
        {du2017deeplog}
\bibfield{author}{\bibinfo{person}{M. Du}, \bibinfo{person}{F. Li}, \bibinfo{person}{G. Zheng}, {and} \bibinfo{person}{V. Srikumar}.} \bibinfo{year}{2017}\natexlab{}.
\newblock \showarticletitle{Deeplog: Anomaly Detection and Diagnosis from System Logs Through Deep Learning}. In \bibinfo{booktitle}{\emph{Proc. of the ACM SIGSAC Conference on Computer and Communications Security}}.
\newblock


\bibitem[Ebert et~al\mbox{.}(2016)]%
        {ebert2016devops}
\bibfield{author}{\bibinfo{person}{C. Ebert}, \bibinfo{person}{G. Gallardo}, \bibinfo{person}{J. Hernantes}, {and} \bibinfo{person}{N. Serrano}.} \bibinfo{year}{2016}\natexlab{}.
\newblock \showarticletitle{DevOps}.
\newblock \bibinfo{journal}{\emph{IEEE software}} (\bibinfo{year}{2016}).
\newblock


\bibitem[Egersdoerfer et~al\mbox{.}(2023)]%
        {10.1145/3588195.3595943}
\bibfield{author}{\bibinfo{person}{C. Egersdoerfer}, \bibinfo{person}{D. Zhang}, {and} \bibinfo{person}{D. Dai}.} \bibinfo{year}{2023}\natexlab{}.
\newblock \showarticletitle{Early Exploration of Using ChatGPT for Log-based Anomaly Detection on Parallel File Systems Logs}. In \bibinfo{booktitle}{\emph{Proc. of the 32nd International Symposium on High-Performance Parallel and Distributed Computing}}.
\newblock


\bibitem[Fu et~al\mbox{.}(2009)]%
        {fu2009execution}
\bibfield{author}{\bibinfo{person}{Q. Fu}, \bibinfo{person}{J.G. Lou}, \bibinfo{person}{Y. Wang}, {and} \bibinfo{person}{J. Li}.} \bibinfo{year}{2009}\natexlab{}.
\newblock \showarticletitle{Execution Anomaly Detection in Distributed Systems Through Unstructured Log Analysis}. In \bibinfo{booktitle}{\emph{Proc. of the IEEE International Conference on Data Mining}}.
\newblock


\bibitem[Ge et~al\mbox{.}(2024)]%
        {ge2024clustering}
\bibfield{author}{\bibinfo{person}{Y. Ge}, \bibinfo{person}{Y. Liu}, \bibinfo{person}{C. Hu}, {et~al\mbox{.}}} \bibinfo{year}{2024}\natexlab{}.
\newblock \showarticletitle{Clustering and Ranking: Diversity-preserved Instruction Selection through Expert-aligned Quality Estimation}. In \bibinfo{booktitle}{\emph{Proc. of the Conference on Empirical Methods in Natural Language Processing}}.
\newblock


\bibitem[Guo et~al\mbox{.}(2024)]%
        {guoowl}
\bibfield{author}{\bibinfo{person}{H. Guo}, \bibinfo{person}{J. Yang}, \bibinfo{person}{J. Liu}, {et~al\mbox{.}}} \bibinfo{year}{2024}\natexlab{}.
\newblock \showarticletitle{OWL: A Large Language Model for IT Operations}.
\newblock  (\bibinfo{year}{2024}).
\newblock


\bibitem[Guo et~al\mbox{.}(2021)]%
        {guo2021logbert}
\bibfield{author}{\bibinfo{person}{H. Guo}, \bibinfo{person}{S. Yuan}, {and} \bibinfo{person}{X. Wu}.} \bibinfo{year}{2021}\natexlab{}.
\newblock \showarticletitle{Logbert: Log Anomaly Detection via Bert}. In \bibinfo{booktitle}{\emph{Proc. of the International Joint Conference on Neural Networks (IJCNN)}}.
\newblock


\bibitem[Gururangan et~al\mbox{.}(2020)]%
        {gururangan2020don}
\bibfield{author}{\bibinfo{person}{S. Gururangan}, \bibinfo{person}{A. Marasovi{\'c}}, \bibinfo{person}{S. Swayamdipta}, {et~al\mbox{.}}} \bibinfo{year}{2020}\natexlab{}.
\newblock \showarticletitle{Don't Stop Pretraining: Adapt Language Models to Domains and Tasks}.
\newblock \bibinfo{journal}{\emph{arXiv preprint arXiv:2004.10964}} (\bibinfo{year}{2020}).
\newblock


\bibitem[He et~al\mbox{.}(2017)]%
        {he2017drain}
\bibfield{author}{\bibinfo{person}{P. He}, \bibinfo{person}{J. Zhu}, \bibinfo{person}{Z. Zheng}, {and} \bibinfo{person}{M.R. Lyu}.} \bibinfo{year}{2017}\natexlab{}.
\newblock \showarticletitle{Drain: An Online Log Parsing Approach with Fixed Depth Tree}. In \bibinfo{booktitle}{\emph{Proc. of the IEEE International Conference on Web Services (ICWS)}}.
\newblock


\bibitem[He et~al\mbox{.}(2020)]%
        {he2020loghub}
\bibfield{author}{\bibinfo{person}{S. He}, \bibinfo{person}{J. Zhu}, \bibinfo{person}{P. He}, {and} \bibinfo{person}{M.R. Lyu}.} \bibinfo{year}{2020}\natexlab{}.
\newblock \showarticletitle{Loghub: A Large Collection of System Log Datasets Towards Automated Log Analytics}.
\newblock \bibinfo{journal}{\emph{arXiv preprint arXiv:2008.06448}} (\bibinfo{year}{2020}).
\newblock


\bibitem[Huo et~al\mbox{.}(2023)]%
        {huo2023semparser}
\bibfield{author}{\bibinfo{person}{Y. Huo}, \bibinfo{person}{Y. Su}, \bibinfo{person}{C. Lee}, {and} \bibinfo{person}{M.R. Lyu}.} \bibinfo{year}{2023}\natexlab{}.
\newblock \showarticletitle{SemParser: A Semantic Parser for Log Analytics}. In \bibinfo{booktitle}{\emph{Proc. of the IEEE/ACM International Conference on Software Engineering (ICSE)}}.
\newblock


\bibitem[Jiang et~al\mbox{.}(2023)]%
        {jiang2023mistral}
\bibfield{author}{\bibinfo{person}{A.Q. Jiang}, \bibinfo{person}{A. Sablayrolles}, \bibinfo{person}{A. Mensch}, {et~al\mbox{.}}} \bibinfo{year}{2023}\natexlab{}.
\newblock \showarticletitle{Mistral 7B}.
\newblock \bibinfo{journal}{\emph{arXiv preprint arXiv:2310.06825}} (\bibinfo{year}{2023}).
\newblock


\bibitem[Jiang et~al\mbox{.}(2024a)]%
        {jiang2024megascale}
\bibfield{author}{\bibinfo{person}{Z. Jiang}, \bibinfo{person}{H. Lin}, \bibinfo{person}{Y. Zhong}, {et~al\mbox{.}}} \bibinfo{year}{2024}\natexlab{a}.
\newblock \showarticletitle{MegaScale: Scaling Large Language Model Training to More Than 10,000 GPUs}.
\newblock \bibinfo{journal}{\emph{arXiv preprint arXiv:2402.15627}} (\bibinfo{year}{2024}).
\newblock


\bibitem[Jiang et~al\mbox{.}(2024b)]%
        {jiang2024lilac}
\bibfield{author}{\bibinfo{person}{Z. Jiang}, \bibinfo{person}{J. Liu}, \bibinfo{person}{Z. Chen}, {et~al\mbox{.}}} \bibinfo{year}{2024}\natexlab{b}.
\newblock \showarticletitle{LILAC: Log Parsing Using LLMs with Adaptive Parsing Cache}.
\newblock \bibinfo{journal}{\emph{Proc. of the ACM on Software Engineering}} (\bibinfo{year}{2024}).
\newblock


\bibitem[Jouppi et~al\mbox{.}(2023)]%
        {Jouppi2023TPUVA}
\bibfield{author}{\bibinfo{person}{N.P. Jouppi}, \bibinfo{person}{G. Kurian}, \bibinfo{person}{S. Li}, {et~al\mbox{.}}} \bibinfo{year}{2023}\natexlab{}.
\newblock \showarticletitle{TPU v4: An Optically Reconfigurable Supercomputer for Machine Learning with Hardware Support for Embeddings}.
\newblock \bibinfo{journal}{\emph{Proc. of the 50th Annual International Symposium on Computer Architecture}} (\bibinfo{year}{2023}).
\newblock


\bibitem[Kim et~al\mbox{.}(2020)]%
        {10.1145/3377813.3381371}
\bibfield{author}{\bibinfo{person}{Jinhan Kim} {et~al\mbox{.}}} \bibinfo{year}{2020}\natexlab{}.
\newblock \showarticletitle{Automatic abnormal log detection by analyzing log history for debugging insights}. In \bibinfo{booktitle}{\emph{Proceedings of the ACM/IEEE 42nd International Conference on Software Engineering: Software Engineering in Practice}}.
\newblock


\bibitem[Le and Zhang(2021)]%
        {le2021log}
\bibfield{author}{\bibinfo{person}{V. Le} {and} \bibinfo{person}{H. Zhang}.} \bibinfo{year}{2021}\natexlab{}.
\newblock \showarticletitle{Log-Based Anomaly Detection Without Log Parsing}. In \bibinfo{booktitle}{\emph{Proc. of the IEEE/ACM International Conference on Automated Software Engineering (ASE)}}.
\newblock


\bibitem[Le and Zhang(2023a)]%
        {le2023log}
\bibfield{author}{\bibinfo{person}{V. Le} {and} \bibinfo{person}{H. Zhang}.} \bibinfo{year}{2023}\natexlab{a}.
\newblock \showarticletitle{Log Parsing: How Far Can ChatGPT Go?}. In \bibinfo{booktitle}{\emph{Proc. of the IEEE/ACM International Conference on Automated Software Engineering (ASE)}}.
\newblock


\bibitem[Le and Zhang(2023b)]%
        {10172786}
\bibfield{author}{\bibinfo{person}{V. Le} {and} \bibinfo{person}{H. Zhang}.} \bibinfo{year}{2023}\natexlab{b}.
\newblock \showarticletitle{Log Parsing with Prompt-based Few-shot Learning}. In \bibinfo{booktitle}{\emph{Proc. of the IEEE/ACM International Conference on Software Engineering (ICSE)}}.
\newblock


\bibitem[Li et~al\mbox{.}(2023)]%
        {li2023did}
\bibfield{author}{\bibinfo{person}{Z. Li}, \bibinfo{person}{C. Luo}, \bibinfo{person}{T.H.P. Chen}, {et~al\mbox{.}}} \bibinfo{year}{2023}\natexlab{}.
\newblock \showarticletitle{Did We Miss Something Important? Studying and Exploring Variable-Aware Log Abstraction}. In \bibinfo{booktitle}{\emph{ICSE 2023}}.
\newblock


\bibitem[Liu et~al\mbox{.}(2024a)]%
        {liu2024loglm}
\bibfield{author}{\bibinfo{person}{Y. Liu}, \bibinfo{person}{Y. Ji}, \bibinfo{person}{S. Tao}, {et~al\mbox{.}}} \bibinfo{year}{2024}\natexlab{a}.
\newblock \showarticletitle{LogLM: From Task-based to Instruction-based Automated Log Analysis}.
\newblock \bibinfo{journal}{\emph{arXiv preprint arXiv:2410.09352}} (\bibinfo{year}{2024}).
\newblock


\bibitem[Liu et~al\mbox{.}(2024b)]%
        {liu2024interpretable}
\bibfield{author}{\bibinfo{person}{Y. Liu}, \bibinfo{person}{S. Tao}, \bibinfo{person}{W. Meng}, {et~al\mbox{.}}} \bibinfo{year}{2024}\natexlab{b}.
\newblock \showarticletitle{Interpretable Online Log Analysis Using Large Language Models with Prompt Strategies}. In \bibinfo{booktitle}{\emph{Proc. of the IEEE/ACM International Conference on Program Comprehension}}.
\newblock


\bibitem[Liu et~al\mbox{.}(2024c)]%
        {liu2024logprompt}
\bibfield{author}{\bibinfo{person}{Y. Liu}, \bibinfo{person}{S. Tao}, \bibinfo{person}{W. Meng}, {et~al\mbox{.}}} \bibinfo{year}{2024}\natexlab{c}.
\newblock \showarticletitle{Logprompt: Prompt Engineering Towards Zero-Shot and Interpretable Log Analysis}. In \bibinfo{booktitle}{\emph{Proc. of the IEEE/ACM International Conference on Software Engineering: Companion Proceedings}}.
\newblock


\bibitem[Luo et~al\mbox{.}(2023)]%
        {luo2023empirical}
\bibfield{author}{\bibinfo{person}{Y. Luo}, \bibinfo{person}{Z. Yang}, \bibinfo{person}{F. Meng}, {et~al\mbox{.}}} \bibinfo{year}{2023}\natexlab{}.
\newblock \showarticletitle{An Empirical Study of Catastrophic Forgetting in Large Language Models During Continual Fine-Tuning}.
\newblock \bibinfo{journal}{\emph{arXiv preprint arXiv:2308.08747}} (\bibinfo{year}{2023}).
\newblock


\bibitem[Ma et~al\mbox{.}(2024)]%
        {ma2024llmparser}
\bibfield{author}{\bibinfo{person}{Z. Ma}, \bibinfo{person}{A.R. Chen}, \bibinfo{person}{D.J. Kim}, {et~al\mbox{.}}} \bibinfo{year}{2024}\natexlab{}.
\newblock \showarticletitle{LLMParser: An Exploratory Study on Using Large Language Models for Log Parsing}. In \bibinfo{booktitle}{\emph{Proc. of the IEEE/ACM International Conference on Software Engineering (ICSE)}}.
\newblock


\bibitem[Makanju et~al\mbox{.}(2009)]%
        {makanju2009clustering}
\bibfield{author}{\bibinfo{person}{A.A.O. Makanju}, \bibinfo{person}{A.N. Zincir-Heywood}, {and} \bibinfo{person}{E.E. Milios}.} \bibinfo{year}{2009}\natexlab{}.
\newblock \showarticletitle{Clustering Event Logs Using Iterative Partitioning}. In \bibinfo{booktitle}{\emph{Proc. of the ACM SIGKDD International Conference on Knowledge Discovery and Data Mining}}.
\newblock


\bibitem[Meng et~al\mbox{.}(2020)]%
        {meng2020logparse}
\bibfield{author}{\bibinfo{person}{W. Meng}, \bibinfo{person}{Y. Liu}, \bibinfo{person}{F. Zaiter}, {et~al\mbox{.}}} \bibinfo{year}{2020}\natexlab{}.
\newblock \showarticletitle{Logparse: Making Log Parsing Adaptive Through Word Classification}. In \bibinfo{booktitle}{\emph{Proc. of the International Conference on Computer Communications and Networks (ICCCN)}}.
\newblock


\bibitem[Meng et~al\mbox{.}(2019)]%
        {meng2019loganomaly}
\bibfield{author}{\bibinfo{person}{W. Meng}, \bibinfo{person}{Y. Liu}, \bibinfo{person}{Y. Zhu}, {et~al\mbox{.}}} \bibinfo{year}{2019}\natexlab{}.
\newblock \showarticletitle{LogAnomaly: Unsupervised Detection of Sequential and Quantitative Anomalies in Unstructured Logs}. In \bibinfo{booktitle}{\emph{IJCAI}}.
\newblock


\bibitem[Messaoudi et~al\mbox{.}(2018)]%
        {messaoudi2018search}
\bibfield{author}{\bibinfo{person}{S. Messaoudi}, \bibinfo{person}{A. Panichella}, \bibinfo{person}{D. Bianculli}, {et~al\mbox{.}}} \bibinfo{year}{2018}\natexlab{}.
\newblock \showarticletitle{A Search-Based Approach for Accurate Identification of Log Message Formats}. In \bibinfo{booktitle}{\emph{Proc. of the IEEE/ACM International Conference on Program Comprehension (ICPC)}}.
\newblock


\bibitem[Narayanan et~al\mbox{.}(2021)]%
        {narayanan2021efficient}
\bibfield{author}{\bibinfo{person}{D. Narayanan}, \bibinfo{person}{M. Shoeybi}, \bibinfo{person}{J. Casper}, {et~al\mbox{.}}} \bibinfo{year}{2021}\natexlab{}.
\newblock \showarticletitle{Efficient Large-Scale Language Model Training on GPU Clusters Using Megatron-LM}. In \bibinfo{booktitle}{\emph{Proc. of the International Conference for High Performance Computing, Networking, Storage and Analysis}}.
\newblock


\bibitem[Oliner and Stearley(2007)]%
        {oliner2007supercomputers}
\bibfield{author}{\bibinfo{person}{A. Oliner} {and} \bibinfo{person}{J. Stearley}.} \bibinfo{year}{2007}\natexlab{}.
\newblock \showarticletitle{What Supercomputers Say: A Study of Five System Logs}. In \bibinfo{booktitle}{\emph{Proc. of the IEEE/IFIP International Conference on Dependable Systems and Networks (DSN)}}.
\newblock


\bibitem[Pan et~al\mbox{.}(2024)]%
        {pan2024raglog}
\bibfield{author}{\bibinfo{person}{J. Pan}, \bibinfo{person}{W.S. Liang}, {and} \bibinfo{person}{Y. Yidi}.} \bibinfo{year}{2024}\natexlab{}.
\newblock \showarticletitle{RAGLog: Log Anomaly Detection Using Retrieval Augmented Generation}. In \bibinfo{booktitle}{\emph{Proc. of the IEEE World Forum on Public Safety Technology (WFPST)}}.
\newblock


\bibitem[Qi et~al\mbox{.}(2023)]%
        {qi2023loggpt}
\bibfield{author}{\bibinfo{person}{J. Qi}, \bibinfo{person}{S. Huang}, \bibinfo{person}{Z. Luan}, {et~al\mbox{.}}} \bibinfo{year}{2023}\natexlab{}.
\newblock \showarticletitle{Loggpt: Exploring chatgpt for log-based anomaly detection}.
\newblock \bibinfo{journal}{\emph{arXiv preprint arXiv:2309.01189}} (\bibinfo{year}{2023}).
\newblock


\bibitem[Rand(1971)]%
        {rand1971objective}
\bibfield{author}{\bibinfo{person}{W.M. Rand}.} \bibinfo{year}{1971}\natexlab{}.
\newblock \showarticletitle{Objective Criteria for the Evaluation of Clustering Methods}.
\newblock \bibinfo{journal}{\emph{J. Amer. Statist. Assoc.}} (\bibinfo{year}{1971}).
\newblock


\bibitem[Sipos and oters(2014)]%
        {sipos2014log}
\bibfield{author}{\bibinfo{person}{Ruben Sipos} {and} \bibinfo{person}{oters}.} \bibinfo{year}{2014}\natexlab{}.
\newblock \showarticletitle{Log-based predictive maintenance}. In \bibinfo{booktitle}{\emph{Proceedings of the 20th ACM SIGKDD international conference on knowledge discovery and data mining}}. \bibinfo{pages}{1867--1876}.
\newblock


\bibitem[Sui et~al\mbox{.}(2023)]%
        {sui2023logkg}
\bibfield{author}{\bibinfo{person}{Y. Sui}, \bibinfo{person}{Y. Zhang}, \bibinfo{person}{J. Sun}, {et~al\mbox{.}}} \bibinfo{year}{2023}\natexlab{}.
\newblock \showarticletitle{LogKG: Log Failure Diagnosis Through Knowledge Graph}.
\newblock \bibinfo{journal}{\emph{IEEE Transactions on Services Computing}} (\bibinfo{year}{2023}).
\newblock


\bibitem[Sun et~al\mbox{.}(2023)]%
        {sun2023text}
\bibfield{author}{\bibinfo{person}{X. Sun}, \bibinfo{person}{X. Li}, \bibinfo{person}{J. Li}, {et~al\mbox{.}}} \bibinfo{year}{2023}\natexlab{}.
\newblock \showarticletitle{Text Classification Via Large Language Models}.
\newblock \bibinfo{journal}{\emph{arXiv preprint arXiv:2305.08377}} (\bibinfo{year}{2023}).
\newblock


\bibitem[Suriadi et~al\mbox{.}(2013)]%
        {suriadi2013root}
\bibfield{author}{\bibinfo{person}{Ouyang Suriadi, Suriadi} {et~al\mbox{.}}} \bibinfo{year}{2013}\natexlab{}.
\newblock \showarticletitle{Root cause analysis with enriched process logs}. In \bibinfo{booktitle}{\emph{Business Process Management Workshops: BPM 2012 International Workshops}}.
\newblock


\bibitem[Tang et~al\mbox{.}(2011)]%
        {tang2011logsig}
\bibfield{author}{\bibinfo{person}{L. Tang}, \bibinfo{person}{T. Li}, {and} \bibinfo{person}{C.S. Perng}.} \bibinfo{year}{2011}\natexlab{}.
\newblock \showarticletitle{LogSig: Generating System Events From Raw Textual Logs}. In \bibinfo{booktitle}{\emph{Proc. of the ACM International Conference on Information and Knowledge Management}}.
\newblock


\bibitem[Tao et~al\mbox{.}(2023)]%
        {biglog}
\bibfield{author}{\bibinfo{person}{S. Tao}, \bibinfo{person}{Y. Liu}, \bibinfo{person}{W. Meng}, {et~al\mbox{.}}} \bibinfo{year}{2023}\natexlab{}.
\newblock \showarticletitle{Biglog: Unsupervised Large-scale Pre-training for a Unified Log Representation}. In \bibinfo{booktitle}{\emph{Proc. of the IEEE/ACM International Symposium on Quality of Service (IWQoS)}}.
\newblock


\bibitem[Tao et~al\mbox{.}(2022)]%
        {tao2022logstamp}
\bibfield{author}{\bibinfo{person}{S. Tao}, \bibinfo{person}{W. Meng}, \bibinfo{person}{Y. Cheng}, {et~al\mbox{.}}} \bibinfo{year}{2022}\natexlab{}.
\newblock \showarticletitle{Logstamp: Automatic Online Log Parsing Based on Sequence Labelling}.
\newblock \bibinfo{journal}{\emph{ACM SIGMETRICS Performance Evaluation Review}} (\bibinfo{year}{2022}).
\newblock


\bibitem[Touvron et~al\mbox{.}(2023a)]%
        {touvron2023llama}
\bibfield{author}{\bibinfo{person}{H. Touvron}, \bibinfo{person}{T. Lavril}, \bibinfo{person}{G. Izacard}, {et~al\mbox{.}}} \bibinfo{year}{2023}\natexlab{a}.
\newblock \showarticletitle{Llama: Open and Efficient Foundation Language Models}.
\newblock \bibinfo{journal}{\emph{arXiv preprint arXiv:2302.13971}} (\bibinfo{year}{2023}).
\newblock


\bibitem[Touvron et~al\mbox{.}(2023b)]%
        {touvron2023llama2}
\bibfield{author}{\bibinfo{person}{H. Touvron}, \bibinfo{person}{L. Martin}, \bibinfo{person}{K. Stone}, {et~al\mbox{.}}} \bibinfo{year}{2023}\natexlab{b}.
\newblock \showarticletitle{Llama 2: Open Foundation and Fine-Tuned Chat Models}.
\newblock \bibinfo{journal}{\emph{arXiv preprint arXiv:2307.09288}} (\bibinfo{year}{2023}).
\newblock


\bibitem[Xie et~al\mbox{.}(2021)]%
        {xie2021logm}
\bibfield{author}{\bibinfo{person}{Yuxia Xie}, \bibinfo{person}{Kai Yang}, {et~al\mbox{.}}} \bibinfo{year}{2021}\natexlab{}.
\newblock \showarticletitle{Logm: Log analysis for multiple components of hadoop platform}.
\newblock \bibinfo{journal}{\emph{IEEE Access}}  \bibinfo{volume}{9} (\bibinfo{year}{2021}), \bibinfo{pages}{73522--73532}.
\newblock


\bibitem[Yang et~al\mbox{.}(2023)]%
        {yang2023baichuan}
\bibfield{author}{\bibinfo{person}{A. Yang}, \bibinfo{person}{B. Xiao}, \bibinfo{person}{B. Wang}, {et~al\mbox{.}}} \bibinfo{year}{2023}\natexlab{}.
\newblock \showarticletitle{Baichuan 2: Open large-scale language models}.
\newblock \bibinfo{journal}{\emph{arXiv preprint arXiv:2309.10305}} (\bibinfo{year}{2023}).
\newblock


\bibitem[Yang et~al\mbox{.}(2024)]%
        {yang2024qwen2}
\bibfield{author}{\bibinfo{person}{A. Yang}, \bibinfo{person}{B. Yang}, \bibinfo{person}{B. Hui}, {et~al\mbox{.}}} \bibinfo{year}{2024}\natexlab{}.
\newblock \showarticletitle{Qwen2 technical report}.
\newblock \bibinfo{journal}{\emph{arXiv preprint arXiv:2407.10671}} (\bibinfo{year}{2024}).
\newblock


\bibitem[Y{\i}ld{\i}z et~al\mbox{.}(2024)]%
        {yildiz2024investigating}
\bibfield{author}{\bibinfo{person}{{\c{C}}. Y{\i}ld{\i}z}, \bibinfo{person}{N.K. Ravichandran}, \bibinfo{person}{P. Punia}, {et~al\mbox{.}}} \bibinfo{year}{2024}\natexlab{}.
\newblock \showarticletitle{Investigating Continual Pretraining in Large Language Models: Insights and Implications}.
\newblock \bibinfo{journal}{\emph{arXiv preprint arXiv:2402.17400}} (\bibinfo{year}{2024}).
\newblock


\bibitem[Zhang et~al\mbox{.}(2007)]%
        {zhang2017syslog}
\bibfield{author}{\bibinfo{person}{S. Zhang}, \bibinfo{person}{W. Meng}, {et~al\mbox{.}}} \bibinfo{year}{2007}\natexlab{}.
\newblock \showarticletitle{Syslog Processing for Switch Failure Diagnosis and Prediction in Datacenter Networks}. In \bibinfo{booktitle}{\emph{Proc. of the IEEE/ACM International Symposium on Quality of Service (IWQoS)}}.
\newblock


\bibitem[Zhang et~al\mbox{.}(2022)]%
        {zhang2022opt}
\bibfield{author}{\bibinfo{person}{S. Zhang}, \bibinfo{person}{S. Roller}, \bibinfo{person}{N. Goyal}, {et~al\mbox{.}}} \bibinfo{year}{2022}\natexlab{}.
\newblock \showarticletitle{Opt: Open Pre-Trained Transformer Language Models}.
\newblock \bibinfo{journal}{\emph{arXiv preprint arXiv:2205.01068}} (\bibinfo{year}{2022}).
\newblock


\bibitem[Zhang et~al\mbox{.}(2019)]%
        {zhang2019robust}
\bibfield{author}{\bibinfo{person}{X. Zhang}, \bibinfo{person}{Y. Xu}, \bibinfo{person}{Q. Lin}, {et~al\mbox{.}}} \bibinfo{year}{2019}\natexlab{}.
\newblock \showarticletitle{Robust Log-Based Anomaly Detection on Unstable Log Data}. In \bibinfo{booktitle}{\emph{Proc. of the ACM Joint Meeting on European Software Engineering Conference and Symposium on the Foundations of Software Engineering}}.
\newblock


\bibitem[Zhang et~al\mbox{.}(2023)]%
        {zhang2023siren}
\bibfield{author}{\bibinfo{person}{Y. Zhang}, \bibinfo{person}{Y. Li}, \bibinfo{person}{L. Cui}, {et~al\mbox{.}}} \bibinfo{year}{2023}\natexlab{}.
\newblock \showarticletitle{Siren's Song in the AI Ocean: A Survey on Hallucination in Large Language Models}.
\newblock \bibinfo{journal}{\emph{arXiv preprint arXiv:2309.01219}} (\bibinfo{year}{2023}).
\newblock


\bibitem[Zhao et~al\mbox{.}(2023)]%
        {zhao2023survey}
\bibfield{author}{\bibinfo{person}{W.X. Zhao}, \bibinfo{person}{K. Zhou}, \bibinfo{person}{J. Li}, {et~al\mbox{.}}} \bibinfo{year}{2023}\natexlab{}.
\newblock \showarticletitle{A Survey of Large Language Models}.
\newblock \bibinfo{journal}{\emph{arXiv preprint arXiv:2303.18223}} (\bibinfo{year}{2023}).
\newblock


\bibitem[Zheng et~al\mbox{.}(2023)]%
        {LogDAPT}
\bibfield{author}{\bibinfo{person}{H. Zheng}, \bibinfo{person}{G. Chu}, \bibinfo{person}{H. Sun}, {et~al\mbox{.}}} \bibinfo{year}{2023}\natexlab{}.
\newblock \showarticletitle{LogDAPT: Log Data Anomaly Detection with Domain-Adaptive Pretraining (industry track)}. In \bibinfo{booktitle}{\emph{Proc. of the 24th International Middleware Conference: Industrial Track}}.
\newblock


\bibitem[Zheng et~al\mbox{.}(2024)]%
        {zheng2024llamafactory}
\bibfield{author}{\bibinfo{person}{Y. Zheng}, \bibinfo{person}{R. Zhang}, \bibinfo{person}{J. Zhang}, {et~al\mbox{.}}} \bibinfo{year}{2024}\natexlab{}.
\newblock \showarticletitle{LlamaFactory: Unified Efficient Fine-Tuning of 100+ Language Models}. In \bibinfo{booktitle}{\emph{Proc. of the Annual Meeting of the Association for Computational Linguistics}}.
\newblock
\urldef\tempurl%
\url{http://arxiv.org/abs/2403.13372}
\showURL{%
\tempurl}


\bibitem[Zhu et~al\mbox{.}(2019)]%
        {zhu2019tools}
\bibfield{author}{\bibinfo{person}{J. Zhu}, \bibinfo{person}{S. He}, \bibinfo{person}{J. Liu}, {et~al\mbox{.}}} \bibinfo{year}{2019}\natexlab{}.
\newblock \showarticletitle{Tools and Benchmarks for Automated Log Parsing}. In \bibinfo{booktitle}{\emph{Proc. of the IEEE/ACM International Conference on Software Engineering: Software Engineering in Practice (ICSE-SEIP)}}.
\newblock


\end{thebibliography}
\end{document}